# 学　士　論　文

題　目　**用例を用いた日本語単語の多義性解消に関する研究**

指導教官　長尾 眞 教授

京都大学工学部 電気工学第二学科

氏　名　松本 光崇

平成6年2月18日

# 目 次

# 第 1 章 序論

自然言語処理研究における重要な課題として単語の多義性の問題がある. 例えば「昼」という単語には「正午」「昼間」「昼食」などの意味がある. このような, 単語の多義性の解消を自動的に行う手法が模索されている. 例えば, 自然言語処理研究の最も直接的な応用分野である機械翻訳においては, 単語の多義性は訳語選択の問題を生じ, 実用的な翻訳技術の確立のために避けて通れない問題である.

言語の曖昧性はこのような意味的曖昧性 (多義性) のみならず, 形態素, 語彙, 構文, 文脈のあらゆるレベルに存在する. これまでにも, それらの個々ないし複数のレベルに対する曖昧性解消の方法が数多く提案されてきたが, 曖昧性解消に共通な問題として次の 2 点が挙げられる. 一つはそれぞれの手法に必要な知識やデータとしてどのようなものを仮定するかという問題, もう一つはそれがどのようにして入手可能かという問題である.

本研究では, 単語の多義性を, 文中に共起する単語を手がかりとして解消することを考えた. 複数の意味概念を持つ単語は, 用いられている文脈から意味概念が決められる. つまり, 単語は表している意味概念によって周囲に共起する単語が異なっていると考えられ, その違いを手がかりに多義性を解消することを試みる. ここで問題となるのは, 多義性解消の手がかりになるであろう共起単語をいかに有効に抽出するか, そして得られた共起単語データをいかに有効に利用するかということである. 本研究では, 単語のそれぞれの語義に対する共起単語データをそれぞれの類義語の用例の中から抽出することを試みた. また構文的に近い位置に共起する単語のみを多義性解消の手がかりとすることによって, 共起単語データの有効的な利用を試みた. このような枠組による実験結果をもとに, より柔軟で有効な多義性解消の方法を考察することも, 本研究の目的の一つである.

以下, 第 2 章では, 共起単語による単語の多義性解消の方法を述べる. ここでは, 多義性の定義, 多義性解消に必要になる共起単語データの抽出の方法, そのデータに基づいた多義性解消の方法について説明する. 第 3 章では, 10 種類の名詞, 計 775 文のテキストに対して行った実験とその実験結果に基づく考察を述べる. 第 4 章では, 本研究の結論を述べる.

# 第 2 章 単語の多義性解消

## 2.1 多義性解消の概要

### 2.1.1 問題の設定

本論文では, ある単語が複数の意味概念 (語義) を持つことを単語の**多義性**といい, そのような単語を**多義語**と呼ぶ. 多義語が用いられる文脈の中で表している意味概念を求めることをもって多義性解消とする. 「昼」という単語の多義性の例を示す.

ぼくは<u>昼</u>を食べてから出かけることにした. (2.1)

この文で「昼」が指すのは「昼食」の意味であり, 英語に訳すなら “lunch” である. これに対して,

彼らは<u>昼</u>も夜もなくただ歩き続けた. (2.2)

ここでの「昼」は「昼間」のことを指し, 英語では “daytime” に相当する.

本研究では, このような単語の多義性解消を, 計算機によって自動的に行うことを試みた. 以下では, その手法および必要になる情報の概略について述べる.

### 2.1.2 多義性解消の概要

単語は文脈によって様々な意味を持って用いられている. 多義性を解消するためには, 単語の各語義に対して, それがどのような文脈で用いられるかについての知識を与えておく必要がある.

本研究では, 文脈の知識を使われている文の中に共起している単語という形で与える. 上の 2 つの文を例に考えてみよう.

例文 (2.1) では, 「昼を食べる」という文脈での「昼」は「昼食」の意味であり, このことは「食べる」という単語によって識別される. これに対して例文 (2.2) では「夜」という単語によって「昼間」という意味の可能性が高いと予想される. すなわち, あらかじめ「昼」という単語の「昼食」の語義に「食べる」「済ませる」などの単語, 「昼間」の語義に「夜」「過ごす」などの単語を共起単語データとし

て与えておけば, 共起単語を調べることにより「昼」が指す意味を決定することができる.

このような立場に基づき, 単語の多義性を解消するまでには, 以下の 3 つの手順が必要になる.

1. 単語の多義性, すなわち単語の語義を定義する
2. 各語義の共起単語データを収集する
3. 得られた共起単語データに基づいて多義性を解消する

以下の節で, これらについて説明をする. まず, 単語の語義の定義は, 本研究では EDR 電子化辞書を用いた. 2.2 節で EDR 電子化辞書の概略, EDR 電子化辞書による語義の定義について述べる. 次に, 共起単語データは, 各語義の類義語が用いられている文を集め, それらの文の中で共起している単語の中から抽出した. 2.3 節で共起単語データの抽出について述べる. 最後に, 得られた共起単語データを有効に利用するための手法を 2.4 節で述べる.

## 2.2 多義性の定義

### 2.2.1 多義性の定義について

多義性解消といったときに, まず必要になるのは語義の定義である. しかし単語の語義の定義の仕方に一般的な見解はなく, 従来の多義性解消の研究では研究者自身が語義を定義し, その定義に基づいて多義性解消が試みられていた. そのため, 研究によって語義の定義が異なっていた.

近年, 従来からある人間が見るための辞書を計算機上で検索できるようにオンライン化した辞書が出てきており, これらの辞書の語義の定義に基づいた研究が行われるようになった. このような標準的な辞書においても, 辞書によって語義の定義や, 語義の分け方の細かさは異なっているが, 客観性があり, 大規模なシステムを目指すためには魅力的な道具といえる.

辞書とは知識や情報を取り出す源である. 多義性解消のためには, 単に語義を定義するだけでなく, それぞれの語義を区別するための情報を取り出さなくてはならない. 現在, 単語のそれぞれの語義の定義文や例文から, 単語やその語義についての知識を抽出する研究が行われている[1)2)]. また, このような自然言語処理研究

のニーズに応えて, 英語では, 計算機による自然言語処理を意識したオンライン辞書の作成がすでに試みられている[3)4)].

日本語においても, 従来からの冊子体の辞書をオンライン化したものが出てきており[5)6)], 計算機による自然言語処理を意識した辞書の開発も進められている[7)8)].

本研究においては多義性の定義に EDR 電子化辞書の単語辞書を用いた. EDR 電子化辞書は, 大規模かつ本格的な計算機処理用の辞書のニーズに応えて開発が始められた辞書で, 今もその開発は進行中である. 規模の大きさ (40 万の日本語単語収録) や情報の多様性を考慮して本辞書を用いることにした.

### 2.2.2 EDR 電子化辞書

EDR 電子化辞書は, 計算機による自然言語処理のために必要な情報を盛り込むことを目的として作成が始められたものである. 構成は, 単語辞書, 概念辞書, 共起辞書, 対訳辞書, EDR コーパスからなり, 現在存在する電子辞書評価版第 2 版には, 単語辞書, 概念記述辞書の一部, 概念体系辞書が含まれている. (図 2.1参照)

単語辞書には, 文の構文構造を把握するための手がかりになる文法的な特性や, その単語が表す概念 (意味) が記述されている. 単語辞書は, 語彙の種類によって, 基本語辞書と専門用語辞書 (情報処理分野) に分けられており, さらに言語の違いによって, 日本語基本語辞書と英語基本語辞書, 日本語専門用語辞書と英語専門用語辞書, に分けられている. 上述のように本研究では単語辞書を語義の定義に用いる.

概念辞書は単語辞書で定義された概念 (意味) についての知識が記述されており, その知識の種類によって概念体系辞書と概念記述辞書に分けられている.

概念記述辞書は, 概念どうしを 18 種類の関係子のリンクによって結びつけた意味ネットワークである. このような意味ネットワークは多義性解消のための有効な知識源になることが考えられるが, 第 2 版では約 530 のリンクが概念体系の上層部で与えられているのみで有効な情報を得ることはできなかった.

概念体系辞書は概念間の階層関係を示したシソーラスである. この体系の中で, 2 つの概念が近い位置にあるということは, それらの概念が類似していることを表している. 本研究では, 語義の情報として類義語を収集する際, また, 単語の意味的な近さを定義する際に概念体系辞書を用いた.

図 2.2 に「昼」を例に概念体系辞書と単語辞書の関係を示す. 単語辞書によって単語「昼」は概念体系辞書中の 3 つの意味概念 (品詞が時詞であるものは除いた)

に対応づけられており, 概念体系辞書から単語が表す各意味概念についての知識が得られる. 図中の矢印は上位概念へのパスを示しており, 特に太矢印は最上位概念への最短パスを示している.

なお次版以降の予定に含まれている共起辞書とは, 単語に文生成のための情報を与えるものであり, 単語の多義性解消に必要な, 語義を識別するための辞書ではない.

### 2.2.3 EDR電子化辞書による多義性の定義

単語辞書のうち日本語単語辞書は約15万5千の単語のデータが収められており, これらの単語は概念辞書における約32万8千の概念と対応付けられている. 単純に計算すると一単語につき2.1個の概念(語義)があることになり, これによって単語の多義性が定義される.

EDR電子化辞書による多義性の定義は主観的に見て細かいことが多く, 2つの語義の違いの判別が難しいことがある. (表3.1 中の単語「隣」の2つの語義など)

## 2.3 共起単語データの抽出

### 2.3.1 概要

多義性の定義に用いたEDR電子化辞書では, それぞれの語義は意味概念を階層構造にまとめたシソーラスの中に位置付けられている (図 2.2 の例). したがって, 各語義に対して, 意味的に類似する語義を持つ単語 (以下**類義語**と呼ぶ. 図 2.3参照) を得ることができる. 本研究ではこの類義語情報を利用して各語義の共起単語データを取り出す.

以下の手順で共起単語データを収集する.

ステップ1：多義語の各語義に対する類義語を取り出す

ステップ2：コーパスから類義語の用例となっている文を取り出す

ステップ3：取り出した文から共起単語データを収集する

この3つのステップによって共起単語データを収集するが, この時, 各ステップを経る間に次第に必要な情報が失われ不要なノイズが入ってくるという問題が生じる. 具体的には, まずステップ1では類義語の中に適当でないものが含まれてい

る場合がある. ステップ 2 では, 類義語として意味的には類似した単語であっても, 用いられ方は全く違う単語である可能性がある. また, 類義語に多義性がある場合は, 類義語が意味的に類似しない語義で用いられている用例が含まれることがある. さらに, ステップ 3 では全ての単語が多義性解消の手がかりになる単語というわけではない.

ステップ 1 のノイズは辞書の問題であり以下では議論はしない. ステップ 2, ステップ 3 におけるノイズについては, それぞれ 2.3.3節, 2.3.4節で議論する.

### 2.3.2 各語義の類義語の収集

共起単語データの作成の第 1 ステップとして, EDR 電子化辞書から各語義の類義語を収集する. 例えば「昼」の場合, 次のような類義語が得られた (図 2.3 参照).

正午の時分： お午 正午 午の時 正時 白昼 真昼 昼なか

昼の食事： 昼食 お昼 昼飯 御飯 飲食物 おやつ

朝から夕方までの間： 昼間 暁 朝 毎朝 朝方

### 2.3.3 類義語の用例の収集

各語義に対する類義語が用いられている例文をコーパスから取りだし, そこで共起している単語を共起単語とすることを考える. 例としてコーパス中に「昼」の語義「昼の食事」の類義語「お昼」を含む次のような文があったとする.

食堂で <u>お昼</u> を食べてから, 授業に出ることにした. (2.3)

この文から語義「昼の食事」の共起単語として「食べる」や「食堂」をはじめ,「授業」「出る」などが得られる. また語義「朝から夕方までの間」の類義語「昼間」を含む次のような文,

施設を利用して, <u>昼間</u> とか夜だけ預かることもできる. (2.4)

がコーパス中にあった場合, 語義「朝から夕方までの間」に対して,「夜」や「預かる」「利用する」などが共起単語として得られる.

類義語自身の多義性によって, 類義語が, 類似していない意味で用いられている例文も得られ, 共起単語データに適当でない共起単語が入ってくることが考えら

れる．この問題については, 多くの種類の類義語の例文を取り出すことによって類似する語義以外の各語義の用例の割合が減少し, そこから得られる共起単語データが, その意味概念のための共起単語データに近似していくと考えた．

### 2.3.4 類義語の例文からの共起単語の収集

上の文 (2.3) で「食べる」という共起単語データが得られるが, 共起単語のうち「授業」や「出る」, また, 文 (2.4) での「施設」や「利用する」という単語は共起単語データとしては必要ではなく, むしろノイズとなる．したがって, 多義性解消の手がかりになる共起単語が, どの位置に現れるかを考える必要がある．このことを考慮した結果, 以下の 4 つの規則に適合する単語だけを共起単語データとして取り出すことにした (下線部が取り出す単語).

1. 類義語が複合名詞の一部となっているとき, 複合名詞を構成する他の単語
   例: 明日の**お昼**過ぎには電話します．

2. 類義語 (または類義語が構成する複合名詞) が係る単語
   例: 食堂で**お昼**を食べてから, 授業に出ることにした．

3. 類義語 (または類義語が構成する複合名詞) が係る単語に係っている他の単語
   例: 食堂で**お昼**を食べてから, 授業に出ることにした．

4. 類義語 (または類義語が構成する複合名詞) に係る単語
   例: あの日食べた**お昼**は最高においしかった．

これらの位置以外に多義性解消の手がかりになる単語が共起することはあり得るが, ノイズを増やさないためにこのように定義した．

1 つの単語が複数回, 共起単語データとして得られる場合は, 得られる回数分だけデータ中に含めることとする．なお上の規則 1 と規則 3 で得られる単語に比べ, 規則 2 と規則 4 から得られる単語の重要度が高いことが多いので, 規則 2 と規則 4 から得られる単語には 2 倍の重みを与えることにした．

## 2.4　共起単語データによる多義性解消の方法

### 2.4.1　キー単語の取り出し

与えられた文中の単語の多義性を, 上記の方法で獲得された共起単語データをもとに解消する.

与えられた文 (以下入力文と呼ぶ) の中に共起している単語にも, 類義語の用例文と同様, 多義性解消に有効な単語とそうでない単語がある. すなわち, 文 (2.1) の「昼」は「食べる」という単語, 文 (2.2) の「昼」は「夜」という単語が多義性解消のキーになる単語 (以下キー単語と呼ぶ) であり, その他の単語はキーにならない. このため, 入力文中の共起単語からキー単語を取り出し, キー単語に基づいて多義性を解消することにする. キー単語の取り出し方としては, 前節で与えた 4 つの規則を用いる.

### 2.4.2　共起単語データとの重複

上述の方法で入力文からキー単語を取り出し, 各語義の共起単語データ中にそれらがどれだけ含まれているか数え, 重複の合計が最も大きい語義を選択することにする.

「昼」の例を用いて具体的に説明する. 「昼」の各語義の共起単語データとして次のようなデータが得られているとする.

正午の時分： 過ぎる　過ぎる　手段　決勝　なる

昼の食事： 食べる　食べる　食堂　おいしい　最高

朝から夕方までの間： 夜　夜　適応する　預かる　過ごす

入力文として次の文が与えられたとする.

一部の生物は体内時計という手段で夜と昼に適応する.　(2.5)

まず, 4 つの規則を適用してキー単語を取り出す. 規則 2 から「適応する」, 規則 3 から「生物」「手段」「夜」がキー単語として得られる (規則 2 で得られたキー単語は重みを 2 倍にすることに注意).

次に得られたキー単語と各語義の共起単語データとの重複の割合を計算する. 「朝から夕方までの間」の語義の場合, 共起単語データの中には 5 個の単語があるので以下のように計算される.

- 「適応する」は1個あるから0.2点.
- 「夜」は2個あるから0.4点である.
- 「適応する」には2倍の重みを与えて0.4点.
- 以上を合計し, 0.8点になる.

同様にして, 語義「正午の時分に」は0.2点, 語義「昼の食事」は0点であり, この結果, 語義「朝から夕方までの間」が選択される.

### 2.4.3 シソーラスの意味的距離を使った多義性解消

キー単語がどの語義の共起単語データ中にも含まれない場合がある. 事実, キー単語となりうる単語を全て共起単語データとして収集することは不可能といってよい. そこでキー単語から意味的に最も近い単語が, 共起単語データに含まれている語義を選択することが考えられる. 次の文を考える.

昼をかき込んで出かけよう. (2.6)

この文から得られるキー単語は「かき込む」のみであり, 共起単語データ中に「食べる」はあるが, 「かき込む」はないとする. 「食べる」と「かき込む」は意味的に近いと考えられ, この場合には意味的近さから多義性が解消されるようにしたい. そこで, シソーラスを用いて単語の意味的近さを定義し, キー単語と共起単語データに対して単に重複を調べるだけでなく, それらの間の意味的近さを調べることにする. シソーラスとしてEDR概念体系辞書(2.2.2節参照)を用いる.

キー単語から意味的に最も近い単語が共起単語データ中に含まれる語義を選択こととし, 同じ距離の単語が複数の語義の共起単語データに含まれる場合には, 前節の重複を求めるときと同様, それらの共起単語データ中での割合が最も大きい語義を選択する.

# 第 3 章 実験と考察

## 3.1 実験

### 3.1.1 実験の準備

#### 多義性解消の対象とする単語の選択

2.2.3 節で例を示したように, EDR 電子化辞書の語義の定義は全体的に細かく, 単語に与えられた語義の違いを人間でも区別できないものが少なくない. 実験対象の単語としては, 語義の区別が明確であり, コーパス中で各語義が万遍なく用いられていることが望ましい. これらのことを考慮して,「足跡」「雨」「キャンプ」「たばこ」「隣」「波」「日」「光」「昼」の 10 単語を実験対象の単語とした. ただし, 文法属性が時詞, 助数詞など, 名詞以外である語義は除外した. 各単語の語義の定義を表 3.1, 表 3.2 に示す.

#### 用例を取り出すコーパス

実験対象とする例文, および共起単語データの抽出に必要な例文を, コーパスから収集する. 実験で用いたコーパスは, 朝日新聞の社説 (1985 年から 1989 年: 約 12 万文), 天声人語 (1985 年から 1991 年: 約 7 万文), 日経サイエンス (1990 年: 約 2 万文) であり, 合計約 21 万文の規模である.

#### 実験の対象とする例文の収集と正解の設定

実験対象として選択した 10 単語が用いられている例文を, コーパスから収集し, 多義性解消の実験対象文とした (計 775 文).

収集した実験対象文の実験対象単語に対して, 正解の語義を人手であらかじめ付与した. 語義の区別がそれほど明らかでない場合や, 周りに共起している単語から正解を一意に設定できない場合は, 正解を二つ以上に設定したものもある.

### 3.1.2 実験

実験の手順を図 3.1 にまとめて示す. 以下その内容を述べる.

#### 類義語の抽出

EDR 電子化辞書から各語義に対して類義語を取り出した. 語義によって, 類義語が 1 つも得られないものもあれば, 1 万語以上も得られるものまであった. 一つ一つの類義語の妥当性を確認するために, 取り出す類義語は 64 語以内に制限した. また類義語が 1 つも得られない語義はここで除外し, その語義で用いられている例文は実験対象の中から除外した. 付録 A に類義語として用いた単語の一覧を示す.

#### 類義語の用例の抽出と共起単語データの抽出

類義語が用いられている例文をコーパスから取り出した. この例文を構文解析して, 2.3.4 節の 4 つの規則に基づき, 共起単語データを収集した. 単語「雨」の各語義の共起単語データを表 3.4に示す.

#### 多義性解消

775 文の実験対象文を構文解析し, 2.3.4 節の 4 つの規則に基づき, キー単語を取り出した. 構文解析, キー単語の抽出に誤りがないか人手で確認し, あれば修正した.

キー単語と得られた共起単語データを用い, 2.4.2 節で述べた重複を数える方法で語義を決定した. この結果を表 3.5 に示す. 次に, 取り出されたキー単語が, いずれの語義の共起単語データにも含まれず語義が決定されなかったものに対し, 2.4.3 節で述べたシソーラス上での意味的近さを比べる方法で語義を決定した. この結果を表 3.6 に示す.

## 3.2 考察

### 3.2.1 国語辞典を用いた予備実験

本実験の前に, 予備実験的な目的で次のような実験を行った. まず, 各語義の類義語 (EDR 電子化辞書で取り出したもの) を, オンライン化されている国語辞典

「大辞林」でひき, その定義文(例文を含む)中に含まれる単語を共起単語データとした. 次に, 実験対象のテキストからは一文中に含まれる全ての単語をキー単語として取り出し, 2.4.2 節で述べた共起単語データ中の単語との重複を数える方法で, 多義性解消を試みた. 結果は, 全体で 45.7% の正解率であった(表 3.3).

この実験では次の 2 点に問題があると考えた.

1. 共起単語データを国語辞典の定義文から取り出したこと.
   定義文は, 単語の指す意味概念を人間が分かるように説明した文で, その中の単語は, 単語が用いられる際に共起する単語とは必ずしも一致しない.

2. 一文中に共起する全ての単語をキー単語としたこと.
   多義性解消の手がかりになる共起単語は, ごく一部の単語だけであり, それ以外の単語が大きなノイズとなって, 多義性解消を妨げた.

この問題点を考慮して, 本論文では, これまで説明してきたように以下の方法を試みた.

1. 共起単語データを用例から取り出す.

2. キー単語は周囲に共起する一部の単語のみを, 2.3.4 節で述べた 4 つの規則を与えて取り出す.

しかしこの方法においても, その枠組は依然として基本的なものであり, 改善の余地が多分にあると思われる. 以下では, 本実験によって得られた結果, 生じた問題点について検討し, それに基づいてより有効な多義性解消の方法を考察する.

### 3.2.2 実験結果についての考察

実験対象のテキストに対し, 2.4.2節で述べた重複を数える方法を適用した結果, 実験対象のテキスト全体で解答が得られた割合は 55.2%, そのうち正解となった割合は 58.2% であった. 次に, 解答が得られなかったテキストに対して, 2.4.3節で述べたシソーラスを用いる方法を適用した結果, 正解率は 39.3% であり, 合計すると全体で 49.7% の正解率であった. なお全体でランダムに語義を選択した場合の正解確率は 37.3% である. 本節では, 重複を数える方法で得られた結果について考察する.

表 3.5 から, 正解率は単語によってばらつきがあることが分かる. 「たばこ」の 87.2%, 「昼」の 79.3%, 「雨」の 75.2% は正解率が高い例であり, 「日」の 12.0%,

「波」の 13.0% は低い例である. 各単語の結果を調べたところ, 正解率の違いは語義の分かれ方が理由の一つであることが分かった. そこで, 以下では実験対象の単語を語義の分かれ方によって分類し, それぞれについて考察を行う. (各単語の語義は表 3.1, 3.2)

なお, 以降「多義性解消の手がかりとなる単語」といったとき, 人間が語義を選択するときに手がかりになる単語を指し, 本実験で収集した共起単語データ中に含まれている単語を指すのではないことに注意しておく.

1. 語義どうしが意味的に遠い単語

   「雨」「昼」「たばこ」がこの例であり, 正解率が高い. 係る動詞や名詞が語義によって異なっていることが多く, 多義性解消の手がかりとして有効である. 係る動詞や名詞は共起単語データとして取り出しやすく, 正解率が高くなる.

   単語「雨」の場合, 語義「雨という天候」は雨が降るという現象を指し, 語義「降ってくる雨水」は雨水という具体物を指しており*, 意味的に遠いものである. 多義性解消には「雨の 日」「雨が 降る」など, 係る名詞や動詞を手がかりにできることが多い.

2. 比喩的な意味の語義を持つ単語

   「足跡」「波」「光」がこの例であり, 正解率は低い. 係る動詞は多義性解消の手がかりにならず, むしろ誤った語義の選択に導く場合がある. 多義性解消の手がかりになる単語は, 前から係ってくる名詞や, 係る動詞の他の格要素であることが多い.

   単語「足跡」の場合, 語義「人や動物が歩いた後に残る跡」に対し, 語義「これまでに残した業績」は比喩的な語義である. 前者の語義の例文「犯人が足跡を残す.」と, 後者の語義の例文「行政に足跡を残す.」を考える. いずれも係る動詞が「残す」で, 多義性解消の手がかりになる単語は「残す」に係る格要素「犯人」「行政」であると思われる. これらの単語は共起単語データ中になく, 共起単語データ中に多く存在する「残す」が語義選択に大きく影響する. 実験では「残す」は前者の語義の共起単語データとして多く取り出

* 「雨」は比喩的な語義「おびただしく降りそそぐもの」を持つが, 実験対象のテキスト中では頻度が低く (165 文中 2 例), 正解率に影響を及ぼさなかった. ここではこの語義を考慮に入れずに議論を進める.

されたので, 比喩的な意味で「足跡を残す」と用いられたテキストはほとんど誤りになった.

3. 語義どうしが意味的に近い単語
「キャンプ」「隣」「歴史」がこの例であり, 正解率が低い. 共起する単語が語義によってあまり違わないため, 多義性解消が困難である. 人間が見ても, 一文中に共起する単語だけでは多義性解消ができないテキストも多い. このような単語について以下で議論はしない.
「キャンプ」の場合, 正解率は高いが, これは複合名詞「難民キャンプ」という形で用いられたテキストが多く, 正解語義の共起単語データ中に単語「難民」が得られていたためである. しかし一般には多義性解消が難しい単語である. すなわち「キャンプ」の語義は「特定の人々を強制的に入れる施設」や「風雨をさけるために幕を支柱に立てて作った小屋」など, いずれも建物で, 建物が使用される目的によって区別されているといえる. したがって, 共起する単語よりも, 文脈によって語義が決まる単語である.

4. 共起単語以外の情報を必要とするもの
「日」は最も正解率が低い. この単語は以下の 2 点の問題がある.
   - 第一に「日本」「日曜日」などの省略語として用いられる語義に対し, 他と同様に類義語の用例から共起単語データを取り出す手法を用いたことが問題である. 語義「日本というアジアにある国」の類義語は「日本」「日本国」などであるが, 意味的に類似していても用いられ方は全く異なる.
   - 第二に, 時詞的に用いられている実験対象テキストについても, 名詞としての語義「限られた一定の期日」を正解に設定して実験を行ったことに問題がある.

   これらの問題は 3.2.5 節で触れる.

### 3.2.3 キー単語を取り出す範囲の拡大

実験では 2.3.4 節で与えた 4 つの規則の位置に共起する単語をキー単語として取り出した. 多義性解消の手がかりとなる単語はこれらの位置以外にも共起することがあり, キー単語を取り出す範囲を拡大する必要がある. 取り出したキー単語

について調べた結果, 次の二つの位置の単語をキー単語に加えることが有効である場合が目立った.

1. 並列句の場合, 構造的に同じ位置の単語
次に示す「昼」の例文中, 4 つの規則によるキー単語は「動かす (重み 2)」「施設」「体」であるが, いずれの単語も多義性解消の手がかりにはならない. 新たな規則で「夜」をキー単語として取り出すことが必要である.

**昼**は施設で体を動かし, 夜 は自宅でぐっすり眠るようになった. (3.1)

2. 係る動詞の他の格要素が名詞句になっている場合, 名詞句中の単語
次の例文中の「足跡」は比喩的な意味で用いられている. 「世界」を手がかりにして決めることも不可能ではないが, 新たな規則によって「老人福祉」あるいは「福祉」をキー単語として取り出せれば, より有効である.

本田さんは老人福祉の世界に大きな**足跡**を残した. (3.2)

キー単語を取り出す範囲を拡大すると, 当然, キー単語中に, 多義性解消の手がかりにならない単語 (ノイズ単語と呼ぶ) が増える. 上に述べた二つの位置にも, ノイズ単語が共起している場合が多い. この問題に対処する方法として, キー単語を順次取り出すことを考えた. 詳しくは 3.2.5 節で述べる.

### 3.2.4 シソーラスを用いる有効性

解答が得られなかったテキストは, 2.4.3 節で述べた方法で, シソーラスを用いて解答を出した. その結果, 解が得られないテキストの一部に正解を与えることができた (表 3.6). しかし本実験のあらゆる単語の組合せで意味的近さを求める方法では, シソーラスを有効に利用できるとはいえない. 例えば次の文で「昼」は語義「朝から夕方までの間」である.

**昼**は妻が子供を迎えに行く. (3.3)

得られるキー単語「行く」「迎え」「妻」「子供」が共起単語データ中になく, シソーラスを用いることになる. 語義「昼の食事」の共起単語データとして, 単語「お代わり」が取り出されている. このとき「妻」と「お代わり」がシソーラス上で意味的に近いため (「妻」の語義「さしみなどのあしらいに使う野菜や海草」と

「お代わり」の語義「一度飲食した後さらにもう一度追加する同じ飲食物」が距離2), この実験文の「昼」に「昼の食事」の語義が選択されてしまう.

このような問題は, 比較する単語を限定することによって, ある程度解決できる. すなわち, 共起単語データ中の単語を, 共起する位置ごとに分類しておき, 構造的に同じ位置の単語どうしのみ意味的近さを比較することをを行えばよい. この方法の詳細は次節で述べる.

### 3.2.5 より有効な多義性解消の方法

これまでに述べてきた考察から導かれる結論として, より有効な多義性解消のための3つの枠組を説明する.

#### 共起単語データの分類とキー単語取り出しの順序の設定

比喩的語義に対する有効なキー単語を取り出し (3.2.2節の2), キー単語の取り出し範囲の拡大によるノイズを防ぎ (3.2.3節), シソーラスを有効に利用 (3.2.4節) するためには, 共起単語データ中の単語をあらかじめ共起する位置によって分類しておき, さらに, キー単語に優先順位を設定して順次取り出していく必要がある. 例えば次の文に対して, 本実験ではキー単語「高い (重み2)」「吸う (重み2)」「草木」を同時に取り出した.

草木はかなり酸性度の高い**雨**を吸っている. (3.4)

しかし, 「雨」が雨水の意味で用いられていることは, 係る動詞「吸う」だけ参照すれば充分である. 他の単語は手がかりにならず, むしろ場合によってはノイズとなり誤った語義の選択を導いてしまう. このことから, 係る動詞「吸う」のみを最初にキー単語として取り出し, 多義性解消の手がかりとすることが有効である.

このように, キー単語の取り出しに順序を設けることが有効であると思われるが, 最適な順序については, 実験的に調べることが必要である. 一例をあげると, おそらく次のような順序が有効であると思われる.

1. 係る動詞. 例文3.4の「吸う」

2. 係る名詞, 形容詞. 「**雨**の日」の「日」

3. 前から係ってくる動詞. 「昨日降った**雨**」の「降る」

4. 前から係ってくる名詞, 形容詞. 例文 3.4 の「高い」

5. 係る単語に係っている他の単語 (係る単語が動詞であれば他の格要素).

6. 5 の単語に係っている単語. 例文 3.2 参照

7. 並列句で構造的に同じ位置の単語. 例文 3.1 参照

この順序でキー単語を取り出し多義性解消する方法を, 文 3.2 を例に説明する. 以下では, 単語「足跡」の共起単語データとして理想的なものを仮定しておく. まず 1 で単語「残す」が取り出される. 「残す」は多義性解消の手がかりにならない単語であるとして 2 に進む. 2, 3 で取り出される単語はない. 4 で「大きい」が取り出されるが手がかりにならない. 5 で「本田さん」「世界」が取り出されるが手がかりにならない. 6 で「老人」「福祉」が取り出され「福祉」によって, 語義が決定される.

### 共起単語以外のデータの必要性

キー単語を上述のような順序で取り出し, 共起単語データを参照するという方法よりも, 文法的情報, 構文的情報がより有効である例を検討する. 3.2.2 節で述べた, 単語「日」の例から示す.

1. 語義に省略語が含まれるもの

   省略語の場合, 係る動詞などは多義性解消の手がかりとならないことが多い. 構文的に並列に共起している単語, あるいは名詞句をなす他の形態素に同様な省略語があるかどうかで決定される.

   〜の量は日, 米, 仏, 英が多い. (3.5)

   この例で, 係る形容詞「多い」は多義性解消の手がかりにならない. 最初に「米」「仏」「英」をキー単語として取り出すことが有効である. この場合, 省略語という文法属性を利用する必要がある.

2. 語義に時間を表すものが含まれるもの

   係る単語, 係ってくる単語, ともに手がかりにならないことが多く, 語義の区別は難しい. 表層格情報や深層格フレームに基づく構文的情報が必要になる.

   春の日を浴びて, 猫がのんきそうに寝ころがっている. (3.6)

ある春の日_彼はのんびりとシャワーを浴びていた. (3.7)

この例では, 係る動詞「浴びる」, 係ってくる名詞「春」は共通であり, 多義性を解消できない. 多義性解消のためには, 前者の「日」は「浴びる」の目的格で, 後者は副詞的に用いられていることを構文的な情報として獲得する必要がある.

### 共起単語データの知識抽出源について

共起単語データを収集する最初の段階として, 類義語をEDR電子化辞書から取り出した. しかし類義語を用いる方法には以下の二つの問題点がある.

1. 意味的に類似していても, 用いられ方が異なる場合がある.
   例えば, 単語「雨」の語義「おびただしく降りそそぐもの」に対する類義語「シャワー」は意味的に類似しているが「弾丸の雨」「涙の雨」という表現に対して, 「弾丸のシャワー」という用例を取り出すことは不可能であろう. このように単語固有の表現で, 類義語からはその共起単語を取り出せない場合がある.

2. 類義語の多義性の問題.
   この問題に対しては, 2.3.4 節で述べたように, 多くの種類の類義語の用例文を取り出すことによって, そこから得られる共起単語データが, その意味概念のための共起単語データに近似していくと考えた. しかし, ノイズの単語が含まれることに変わりなく, また類義語のコーパス中での出現頻度の偏りや, 語義が用いられる頻度の偏りによってノイズ単語の割合が大きくなる場合がある.

より厳密な共起単語データを収集するためには, 類義語の用例を用いる手法では限界がある. 別の形で, あるいは別の知識源から抽出する必要がある. 以下に考えられるものを二つ述べる.

1. コーパスからの半自動抽出
   コーパス中の多義語に対して人間が正しい語義を与え, それをもとに計算機が自動的に共起単語データを抽出するシステムが考えられる. すなわち, 大規模なコーパスの中から用例を集め, 単語がどの語義で用いられているかを人間が判断して解答を与える. 計算機は文中の単語を共起する位置ごとに分類

し, 与えられた解答をもとに共起単語データとして蓄える. このようなシステムにより, コーパスから半自動的に共起単語を抽出することが可能になる.

2. EDR 概念記述辞書からの抽出

   2.2.2 節で述べた通り, EDR 概念記述辞書は, 概念どうしの関係を 18 種類の関係子で記述した辞書である. 現在の第 2 版では記述の数が少なく知識を抽出するには不十分であったが, 詳細な概念記述が与えられれば, 将来的には次のような共起単語データを抽出することができる.

   例えば, 「昼, 雨が降った.」という文で, 「昼」が語義「朝から夕方までの間」, 「雨」が語義「降ってくる雨の水」, 「降る」が語義「高い位置から低い所へ移り動く」であるとする. このことは, 関係子「agent: 動作を引き起こす主体」と「time: 事象の起こる時間」による次のような概念記述があれば決定できる.

   「高い位置から低い所へ移り動く」–agent→ 「降ってくる雨の水」
   「高い位置から低い所へ移り動く」–time→ 「朝から夕方までの間」

# 第 4 章 結論

本研究では, 共起単語を手がかりとして, 単語の多義性解消を試みた. まず, 単語の各語義の共起単語データを, それぞれの類義語の用例をコーパスから取り出すことによって収集した. 次に, 多義性解消を行う単語に対して, 同一文中に存在する単語の一部を多義性解消のためのキー単語として取り出し, それらを共起単語データと比較することによって語義を選択した. 実験の結果, 語義どうしが意味的に遠い単語については高い正解率が得られた. しかし, 語義どうしが意味的に近い単語や, 比喩的な意味で用いられる語義を持つ単語については, 高い正解率は得られなかった.

この実験結果を検討した結果, 全体の正解率, 中でも比喩的意味の語義を持つ単語の正解率を上げるために, 多義性解消のためのキー単語を順次取り出していく必要があることが分かった. また単語によっては, その文法的属性や, 単語が用いられている文の構文的情報を必要とすることが分かった. このような共起単語データ利用の方法や, 文法的, 構文的情報の利用の方法によって, より有効な多義性解消が可能になることが考えられる.

共起単語データを収集するにあたっては, 本実験で採用した, 類義語を手がかりとする手法には限界がある. 今後, より厳密な共起単語データ抽出のためのシステムや知識源の模索, 作成が必要である.

# 謝 辞

本研究を進めるにあたり、終始懇切丁寧に御指導下さいました長尾眞教授に心から御礼申し上げます。

本研究を進めるにあたり、終始助言および指導をして下さいました黒橋禎夫助手に心から感謝いたします。

本研究を進めるにあたり適切な御指示をいただきました宇津呂武仁氏に心から感謝いたします。

本研究およびプログラミングに関して協力して下さった長尾研究室の皆様に感謝します。

# 参 考 文 献

# 付録 A 類義語

本実験において共起単語データの取り出しのために用いた類義語の一覧を示す。

取り出す類義語は 64 語以内に制限し、さらに多義性などにより共起単語データの取り出しの妨げになるであろう類義語は人手で削除した。

表 A.1: 単語「雨」の類義語

| | |
|---|---|
| 雨という天候 | 秋の空 悪天 雨脚 雨足 雨け 雨降り 暗雲 雨意 雨天 気候 雲間 雲夜 曇がち 曇りがち 曇り勝ち 曇勝 曇勝ち 曇り空 曇空 荒天 小春 小春日 好天 上天気 天気 晴天 積陰 雪意 お天気 ウェザー 気節 空もよう 空合 空合い 空色 空模様 御天気 多雨 多照 梅雨 梅雨時 梅雨入 梅雨入り 梅雨空 天候 日当たり 吹き回し 雪気 雪空 雪ふり 雪降 雪降り 浦和 乞食 行楽日和 長春 うす曇 うす曇り 薄曇 薄曇り |
| おびただしく降り | シャワー 雲煙 雲水 煙火 烟火 影 火水 涓塵 水火 そそぐ物 地気 電気 塗炭 氷炭 放射能 |
| 降ってきた雨の水 | 雨水 降水 天水 いわし雲 さば雲 鰯雲 巻積雲 絹積雲 鯖雲 土埃 雨つぶ 雨粒 玉垂 玉垂れ 雲煙 雲霞 雲霧 煙霞 煙霧 お下がり お下り 御下がり 御下り 風雲 晴天 風雨 風雪 寒空 寒天 冬空 雪 雲雨 スカイライン フォグ フォッグ 霓 光象 虹 虹霓 靄 フォグ フォッグ 霞 晴嵐 晴嵐 薄霧 霧 濛気 |

表 A.2: 単語「足跡」の類義語

| | |
|---|---|
| 通っていった道筋 | 足どり　足取　足取り　家路　往復　往覆　風道　片道　火道　魚道　雲路　血路　黄道　潮筋　潮道　潮路　舟路　条坊　ストレッチ　征途　祖道　近道　近路　長途　波路　浪路　逃げ道　逃げ路　逃道　逃路　逕路　ウェー　ウェイ　ウエー　ウエイ　バーン　ライン　往還　往来　街道筋　街路　径路　車道　車路　小みち　小径　小道　小路　樵路　衢　通い路　通り道　通り路　通道　通路　徒路　途　道筋　道塗 |
| これまでに残した業績 | 家がら　家柄　遺芳　余栄　威霊　烏兎　歳月　歳霜　春秋　星　星霜　年紀　年記　年月　お里　御里　生い立ち　生立ち　画業　学歴　画歴　きず口　傷ぐち　傷口　前非　疵ぐち　疵口　苦汁　句歴　史伝　職歴　種姓　素姓　素性　素生　前身　前歴　年歴　浮生　前　略歴　生活史　前科　金字塔　人生行路　世　代　一代　かた書き　肩がき　肩書　肩書き　肩書　肩書き　ライフサイクル　生活史　キャリア　キャリヤ　キャリヤー　過去　経歴　履歴　ライフ　一世 一生 |
| 人や動物が歩いた後に残る跡 | 迹　跡　足形　足　跡かた　跡形　遺芳　雨痕　お手　御手　男手　軌　軌跡　古轍　故轍　轍　軌跡　括　括れ　縊　縊れ　シュプール　蹤跡　縦迹　人跡　前轍　踪跡　弾痕　糊目　歯形　刷毛目　剝げ　食み　針目　引き目　引目　墓目　焼け焦げ　足取　足取り　打ち目　打目　刷り目　刷目　摺り目　摺目　擦れ目　擦目　ウォーター　マーク　キスマーク　鳥跡　血痕　指紋　形迹　形跡　焦げ目　焦目　飛跡 |

表 A.3: 単語「キャンプ」の類義語

| | |
|---|---|
| 特定の人々を強制的に入れる施設 | 強制収容所　収容所 |
| 風雨をさけるために幕を支柱に立てて作った小屋 | テント　天幕　幕屋　穹廬 |
| 合宿で合同練習をする場所 | 射場　道場　ブルペン　馬場　射場　練兵場 |
| 兵士が集団で居住する設備のある所 | バラック　営舎　営所　宿営　隊舎　兵営　兵舎 |

表 A.4: 単語「たばこ」の類義語

| | |
|---|---|
| 煙草という嗜好品 | 思い草　思草　烟草　金口　シガー　巻きたばこ　巻きタバコ　巻き煙草　巻タバコ　巻煙草　目覚まし草　目覚草　粗葉　シガーレット　シガレット　煙太　紙巻　紙巻き　紙巻きタバコ　紙巻き煙草　紙巻煙草　目ざまし草　目覚し草　烟草　フィルターチップ　ジョイント　きざみタバコ　刻みたばこ　刻みタバコ　刻み煙草　刻煙草　かみタバコ　嚙みタバコ　嚙み煙草　嚙煙草　嗅ぎタバコ　嗅ぎ煙草　嗅タバコ　嗅煙草 |
| タバコという植物 | 烟草　タイマ　大麻　天草　コーヒーの木　コーヒーノキ |

表 A.5: 単語「隣」の類義語

| | |
|---|---|
| 並んで建っている<br>両横の家 | あき家　空き屋　空き家　空屋　空家　明き家　明家　あき巣　空き巣<br>空巣　明き巣　明巣　家々　家家　家並　家並み　庵　廬　板屋　板家<br>一家　一軒　一戸　一軒家　一軒家　丸屋　田舎家　田舎家　田舎家<br>犬小屋　忌み屋　忌屋　岩室　石室　岩屋　窟　石屋　磐屋　隠居<br>うさぎ小屋　家　内　陋居　あばら屋　あばら家　ねぐら　わが家<br>家方　我が家　我家　茅舎　寓　荒屋　荒家　私家　私宅　自家　拙家<br>草屋　草堂　草廬　内方 |
| ある場所の近辺 | ぐるり　界隈　外縁　居回　居回り　境界　近間　近所　近辺　近傍<br>四囲　四顧　四壁　周縁　巡り　附近　辺り近処　辺り近所　辺近処<br>辺近所　町内　お手もと　お手許　お手元　お側　お傍　外囲　居回<br>居回り　近め　近目　御手許　御手元　御側　根際　最寄　最寄り<br>四境　四周　四隣　至近　至近距離　周回　足下　足許　不離　付近<br>片わき　片脇 シャワー　雲煙　雲水　煙火　烟火　影　火水　涓塵<br>水火　地気　電気　塗炭　氷炭　放射能　近隣　手ぢか　隣りあわせ<br>隣り合せ　隣り合わせ　隣合　隣合せ　眼路　目先　目前　直近　臨港<br>目近　近さ　手近さ　間ぢかさ　間近さ　目と鼻の間　指顧　手近<br>手ぢか　あたり近所　まぎわ　間際　近場　根際　手先　真際　足下<br>足許　近め　手近　辺り近処　辺り近所　辺近処　辺近所　近さ |

表 A.6: 単語「波」の類義語

| | |
|---|---|
| 媒質中を振動が伝わってゆく現象 | うねり　ウェーブ　ウェーヴ　ウエーブ　ウエーヴ　波状　波動　アーク放電　アーチ効果　圧電効果　唸り　核分裂　核分裂作用　原子核分裂　後座　コロナ　歳差　歳差運動　磁気　磁気飽和　摂動　電気伝導　電弧　電磁偏向　トランスミッション　内部摩擦　波動　パイロ電気　火花放電　ピロ電気　分極　分散　コロナ放電　ショートサーキット　トンネル効果　$\beta$ 崩壊　マイスナー効果　重力レンズ　ブロッホライン　ファラデー回転　物理現象 |
| 波浪 | 海波　巨浪　大浪　波浪　あだ波　徒波　徒浪　男波　男浪　回瀾　逆浪　逆浪　紅潮　水紋　蒼波　高波　縦波　怒涛　波残り　余波　塩花　波の穂　万波　表面波　風浪　櫓あし　櫓脚　波穂　巨涛　鯨波　洪涛　大波　涛波　波涛　波濤　水波　濁波　女波　海嘯　白波　白浪　さざ波　細波　小波　漣　波跡　逆波　逆浪　荒波　荒浪　静振　横波　うねりシャワー　雲煙　雲水　煙火　烟火　影　火水　涓塵　水火　地気　電気　塗炭　氷炭　放射能 |
| 時代や物事の趨勢 | 動向 |
| ペルシャという，南西アジアにあった国 | ペルシャ　ペルシア |

表 A.7: 単語「日」の類義語

| | |
|---|---|
| 週の一番初めの日 | 日曜　日曜日 |
| 日本というアジアにある国 | ジャパン　日本　日本国 |
| 太陽の光 | 赤日　昼光　天日　日の目　日光　白色光　陽光　月明　ムーンライト　月あかり　月桂　月光　月色　スターライト　星あかり　星影　星光　星芒 |
| 限られた一定の期日 | あくる日　在りし日　五十日　何日　幾日　一日　月旦　月立　朔　朔日　ひと日　某日　一両日　今明　今明日　一六　五日　五日　いにしえ　亥の子　忌み日　海開き　永日 |
| 朝から日暮までの間 | 昼　昼なか　昼ま　昼間　昼中　日なか　日の中　日脚　日足　日中 |
| 太陽という天体 | お天道様　お日さま　お日様　ソレイユ　炎精　火輪　金烏　御天道様　御日様　今日様　赤烏　太陽　天道　天日　日華　日天　日天子　日輪　陽 |

表 A.8: 単語「光」の類義語

| | |
|---|---|
| 日や月などが発する光 | 光　月明　赤日　昼光　天日　日の目　日光　白色光　陽光　来迎　ムーンライト　月あかり　月影　月花　月桂　月光　月色　スターライト　星あかり　星影　星光　星彩　星芒 |
| 心の中にやどる明るさや希望 | 厭　厭き　飽き　呆気　鬱悶　怯え　脅え　恩波　買い気　買気　快哉　感気　危機感　気先　気組み　気篭み　喜色　苦患　苦業　拷問　五苦　寒け　寒気　四苦　辛酸　情合　情合い　水火　酔興　当て事　当事　泣き　辱　恥　羞　恥じらい　半焼　パトス　引け色　引色　不念　無念　慢気　未練　迷霧　老苦　喜憂　三楽　引っかかり　引っ掛かり　引っ掛り　引掛　引掛かり　引掛り　シャワー　雲煙　雲水　煙火　烟火　影　火水　涓塵　水火　地気　電気　塗炭　氷炭　放射能 |

表 A.9: 単語「昼」の類義語

| | |
|---|---|
| 朝から夕方までの間 | 昼間　昼ま　朝 |
| 昼の食事 | お昼　ランチ　ランチョン　午餐　午飯　御昼　中食　昼餉　昼げ　昼御飯　昼餐　昼食　昼飯　昼弁当　朝餉　うまうま　飲食物　餌ば　食餌　食品　扶持　塩噌　追いだき　大炊　お代り　お代わり　御代り　御代わり　お茶の子　御茶の子　御斎　お斎　御斎　おまんま　お飯　御飯　お目覚　御目覚　おやつ　お八つ　御八つ　下物　間食　学校給食　機内食　給食　供御　供御　スナック　小漬け飯　小漬飯　小付け飯　小付飯　小昼　御膳　三食　三食　七五三　主食　常食　膳 シャワー　雲煙　雲水　煙火　烟火　影　火水　涓塵　水火　地気　電気　塗炭　氷炭　放射能 |
| 正午の時分 | お午　まっ昼間　午の時　真昼　昼つ方　昼どき　昼なか　昼頃　昼時分　昼日なか　日ざかり　白昼　明け六つ　朝飯前　亥　位相　一時間　一刻　一点　五つ　乙夜　戌　卯　丑　丑三つ　日暮　薄暮　大引け　元期　際目　計時　鶏鳴　下刻　刻限　刻々　此れ　九つ　五更　五更　三時　潮合　四更　視時　正午　正時　時刻　時日　数刻　辰　知死期　定刻　定時　寅 |

表 A.10: 単語「歴史」の類義語

| | |
|---|---|
| 人間社会の歴史などに関する研究を行う学問 | 歴史学　異学　医用電子工学　英学　英文学　ＳＥ　システム工学　エレクトロニクス　エレクトロニクス　エレクトロニックス　電子工学　会計学　解析　解剖学　歌学　漢学　看護学　漢方　画学　学事　眼科　基礎医学　教育学　教育工学　近代経済学　軍学　ポリティカルエコノミー　経済学　経済史　メタフィジック　形而上学　純正哲学　系統分類学　計量経済学　アーキテクチャ　アーキテクチャー　アーキテクチュア　建築学　言語心理学　工科　工業化学　アーキオロジー　考古学　構造力学　古学　国語学　国語史　国文学　語学　五行　語論　サイエンス　自然科学　理学　史学　詩学　自然科学　自然哲学 |
| 人や組織,事物が現在までにたどった過程 | 哀史　経済史　県史　現代史　校史　国語史　エチモロジー　エティモロジー　語原　語源　語歴　詩史　社史　社歴　史要　事歴　世界史　前史　前史　前史　全史　村史　地史　地方史　通史　秘史　秘史　文学史　新史　近代史　国史　古史　由緒シャワー　雲煙　雲水　煙火　烟火　影　火水　涓塵　水火　地気　電気　塗炭　氷炭　放射能 |
| 事項を起こった年月の順に書いた | クロニクル　メモリアル　縁起状　史学　史冊　史書　青史　年代記　年暦　年歴　編年史　歴代　紀要　署名　自讃　自賛　歴史　断想　補記　補訂　要義　要項　インストラクショインストラクション　上申書　情景描写　キャプション　サイドタイトル　タイトル　字幕　ペーパ　ペーパー　ペープル　ペイパー　論文 |

# 付録 B 多義性解消の手がかりになる単語と共起する位置

単語「雨」と単語「足跡」について, 多義性解消の手がかりになる単語を人手で取り出し, 共起する位置によって分類してみた. 「雨」は, 係る動詞や名詞が手がかりになることが多く、「足跡」は「偉大な**足跡**」など係ってくる単語が手がかりとなることが多い (3.3.2 節参照).

表 B.1: 単語「足跡」の多義性解消の手がかりになる共起単語

| 語義 | 係る動詞 | 係る名詞・形容詞 | 係ってくる単語 | 他の格要素 | 複合名詞 |
|---|---|---|---|---|---|
| 通っていった道筋 | | | | | 追及 |
| これまでに残した | しのぶ 学ぶ<br>理解する 思う | | 偉大な<br>ひたむきな<br>着実な<br>行政<br>戦後社会<br>列強 | 生き方 | |
| 人や動物が歩いた後に残る跡 | つく<br>判別する<br>記録する | 図 | ネズミ<br>昆虫<br>ウサギ | | 採取 |
| (手がかりにならない) | 残す 頼る<br>振り返る さらす<br>整理する 迫る<br>重なり合う | | 先人 | | |

表 B.2: 単語「雨」の多義性解消の手がかりになる共起単語

| 語義 | 係る動詞 | 係る名詞・形容詞 | 係ってくる単語 | 他の格要素 | 複合名詞 |
|---|---|---|---|---|---|
| 雨という天候 | 続く 知る<br>負ける 待つ<br>聞こえる たたる<br>繰り返す 来る | 気配 夜<br>国 季節<br>あと 日<br>日 | 夏<br>激しい<br>果てしない<br>時候 | 時期<br>全土<br>毎度 | |
| おびただしく降り注ぐもの | | | 血 火 弾丸 | | |
| 降ってきた雨の水 | 降る 流す 打つ<br>降り注ぐ 注ぐ<br>降りしきる 頼る<br>濡れる 捨てる<br>利用する よける<br>にじむ 溶ける<br>上がる 蓄える<br>染み込む 染み通る<br>こぼれる 止む<br>吸う | 貴い<br>量<br>モスクワ<br>音<br>一粒<br>半分<br>受け皿 | 降る<br>酸性<br>冷たい<br>けぶる | 川 | 不足 |

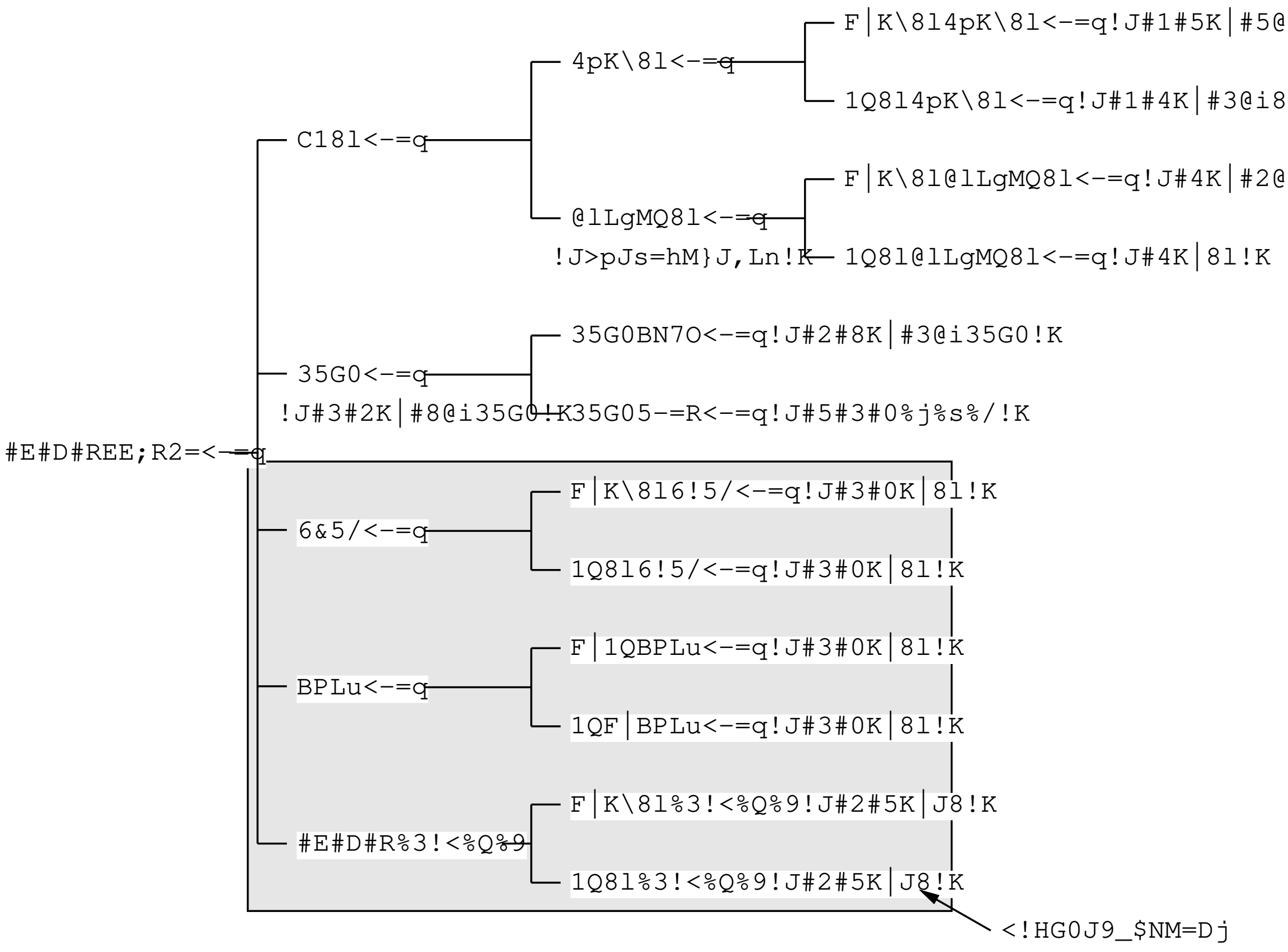

単語辞書 : 文の構文構造を把握するための手がかりになる文法的特性、およびその単語が表す概念を記述した辞書。

概念辞書 : 単語によって表される意味内容・概念の知識を記述した辞書。

共起辞書 : 言葉の言い回しに関する情報を記述した辞書。

対訳辞書 : 日本語と英語の単語見出し間の対応関係の情報を記述した辞書。

EDR コーパス : 大量の用例を収集し、解析して得られる言語データ。

現在の電子化辞書評価版に含まれるのは、単語辞書、概念辞書 (概念記述辞書は一部のみ) である。共起辞書、対訳辞書、EDR コーパスは次版以降での予定に含まれている。

図 2.1: EDR 第 2 版の構成

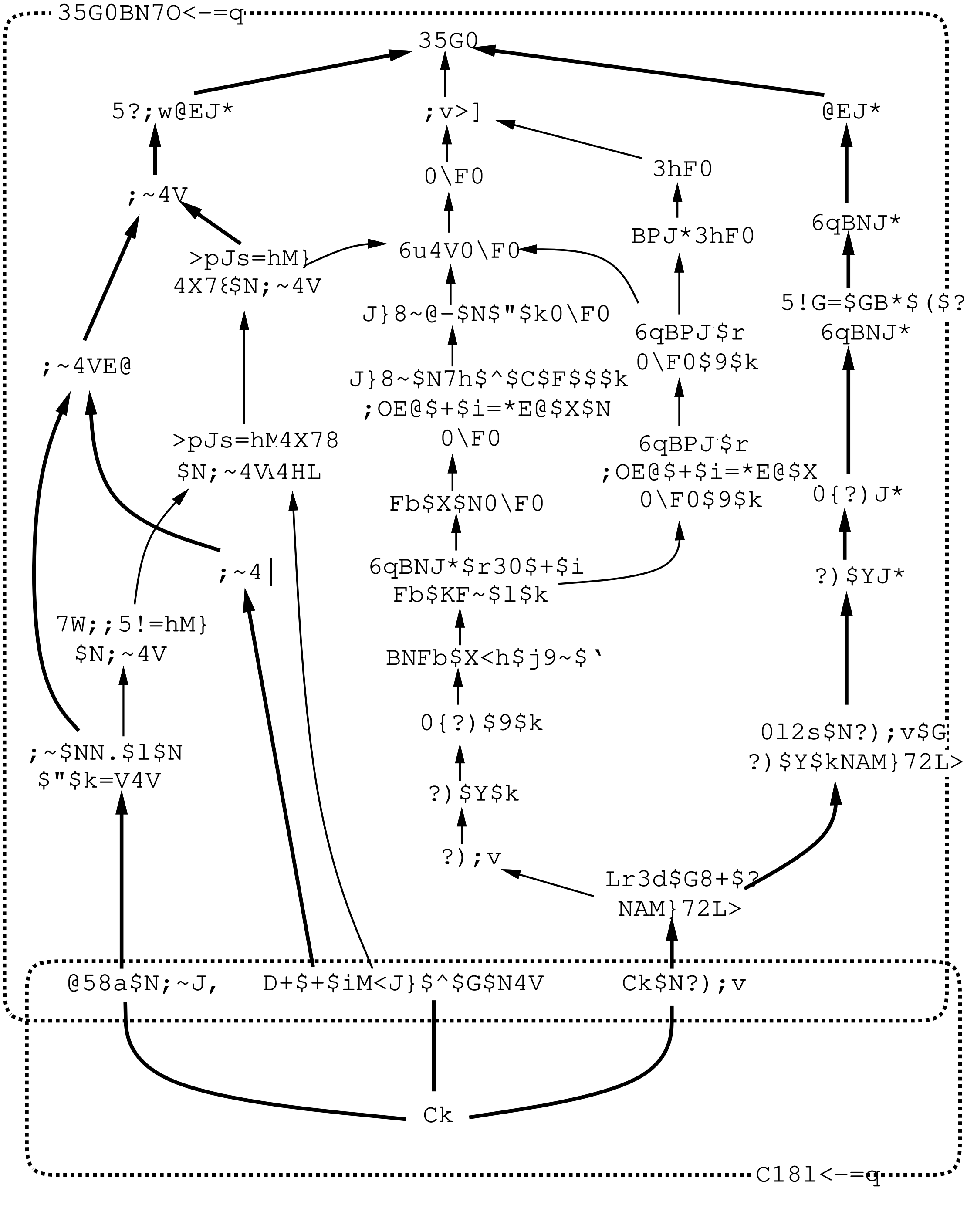

図 2.2: EDR 単語辞書と EDR 概念体系辞書の関係図

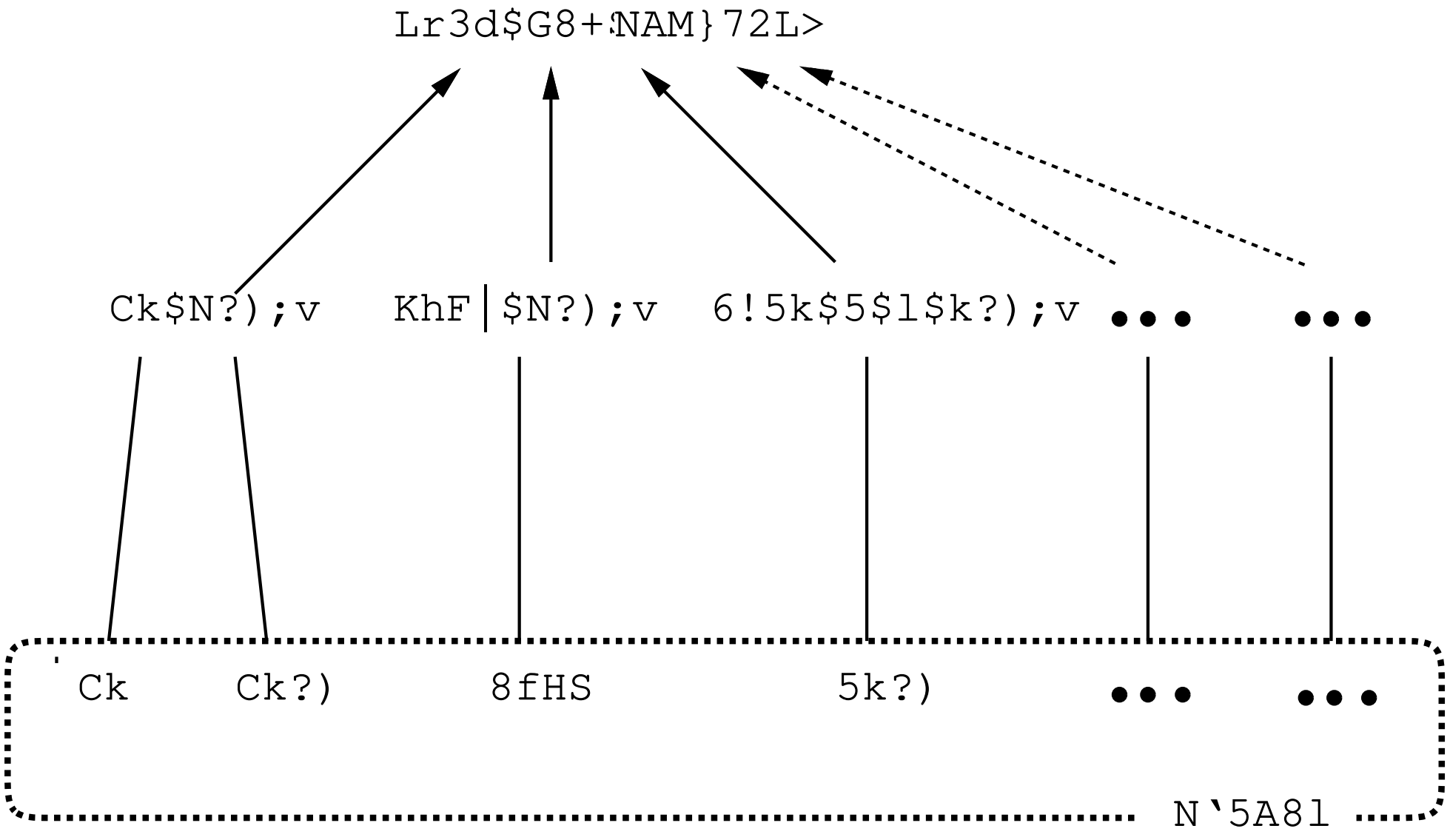

図 2.3: 単語「昼」の語義「昼の食事」に対する類義語

表 3.1: 単語の語義 その 1

| 単語 | 語義 | コメント |
|---|---|---|
| 足跡 | 通っていった道筋 | |
| | これまでに残した業績 | |
| | 人や動物が歩いた後に残る跡 | |
| | (足跡) | 類義語が取り出せない |
| 雨 | 雨という天候 | |
| | おびただしく降りそそぐ物 | |
| | 降ってきた雨の水 | |
| キャンプ | 特定の人々を強制的に入れる施設 | |
| | 風雨をさけるために幕を支柱に立てて作った小屋 | |
| | 合宿で合同練習をする場所 | |
| | 兵士が集団で居住する設備のある所 | |
| | (練習のための合宿) | 類義語を取り出せない |
| たばこ | 煙草という嗜好品 | |
| | タバコという植物 | |
| | (紙巻き煙草) | 意味カテゴリが「煙草と…」と同じ |
| 隣 | 並んで建っている両横の家 | |
| | ある場所の近辺 | |
| | (すぐ近いあたり) | 「ある場所…」との違いが区別できない |
| 波 | 媒質中を振動が伝わってゆく現象 | |
| | 波浪 | |
| | 時代や物事の趨勢 | |
| | ペルシャという，南西アジアにあった国 | |
| | 流動するもの | 類義語を取り出せない |
| | (波のように起伏しつつ揺れ動くもの) | 類義語が意味的に類似していない |
| | (物が連続的に起伏して見える状態) | 類義語が意味的に類似していない |
| | (激しい勢いでよって来るもの) | 類義語が意味的に類似していない |
| | (ポーランドという国) | 意味カテゴリが「ペルシャ…」と同じ |

注) コメント欄に記述がある語義はその理由により実験対象から除外した

表 3.2: 単語の語義 その 2

| 単語 | 語義 | コメント |
|---|---|---|
| 日 | 週の一番初めの日 | |
| | 日本というアジアにある国 | |
| | 太陽の光 | |
| | 限られた一定の期日 | |
| | 朝から日暮までの間 | |
| | 太陽という天体 | |
| | (日という，時間のメートル法単位) | 類義語を取り出せない |
| | (日という，時間の単位) | 類義語を取り出せない |
| 光 | 日や月などが発する光 | |
| | 心の中にやどる明るさや希望 | |
| | (光市という山口県にある市) | 固有名詞である |
| 昼 | 昼の食事 | |
| | 朝から夕方までの間 | |
| | 正午の時分 | |
| | (正午に) | 時詞である |
| | (朝から日暮までの間) | 時詞である |
| 歴史 | 人間社会の歴史などに関する研究を行う学問 | |
| | 人や組織，事物が現在までにたどった過程 | |
| | 事項を起こった年月の順に書いた歴史 | |

注) コメント欄に記述がある語義はその理由により実験対象から除外した

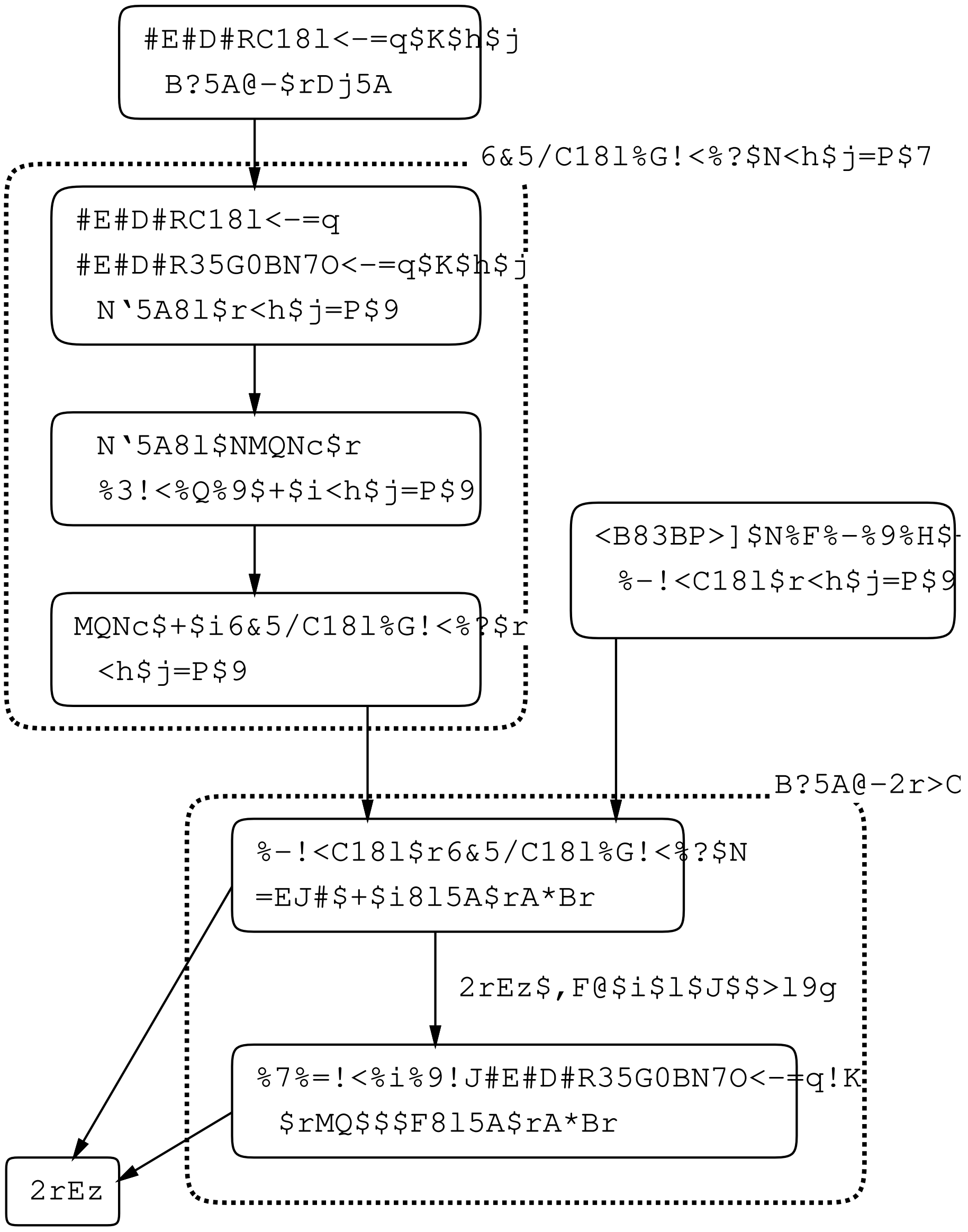

図 3.1: 多義性解消のフローチャート

表 3.3: 「大辞林」の定義文による多義性解消の結果

| | 語義数 | 実験文数 | 正解 | 誤り | 解答なし | 解答率 | 正解率 |
|---|---|---|---|---|---|---|---|
| 足跡 | 3 | 35 | 22 | 12 | 1 | 98.1 | 64.7 |
| 雨 | 3 | 165 | 59 | 101 | 5 | 97.0 | 36.9 |
| キャンプ | 4 | 30 | 6 | 20 | 4 | 86.7 | 23.1 |
| たばこ | 2 | 94 | 57 | 32 | 5 | 94.7 | 64.0 |
| 隣 | 2 | 51 | 35 | 12 | 4 | 92.2 | 74.5 |
| 波 | 4 | 39 | 8 | 31 | 0 | 100.0 | 20.5 |
| 昼 | 3 | 36 | 13 | 23 | 0 | 100.0 | 36.1 |
| 計 | - | 549 | 235 | 279 | 35 | 93.6 | 45.7 |

表 3.4: 単語「雨」に対して抽出された共起単語データ

| 雨という天候 | | 降り注ぐもの | | 降ってくる雨水 | |
|---|---|---|---|---|---|
| 単語 | 数 | 単語 | 数 | 単語 | 数 |
| する | 10 | 落とす | 48 | ためる | 19 |
| 続く | 10 | 薄い | 38 | 降る | 14 |
| 明ける | 8 | 暗い | 25 | 落ちる | 13 |
| ない | 6 | ない | 15 | ない | 11 |
| 悪い | 6 | なる | 13 | 深い | 11 |
| 温暖 | 6 | ある | 12 | 利用する | 10 |
| 夏 | 5 | ひそめる | 12 | する | 9 |
| 変化 | 5 | 濃い | 11 | おおう | 8 |
| 日 | 4 | いう | 9 | なる | 8 |
| 影響する | 4 | 部分 | 9 | 包む | 8 |
| 晴れる | 4 | する | 8 | こと | 7 |
| こと | 4 | つきまとう | 8 | しみこむ | 7 |
| 広がる | 4 | 投げる | 8 | ロンドン | 7 |
| 降る | 4 | ガン | 7 | 使う | 7 |
| 各地 | 4 | 強い | 7 | まかなう | 7 |
| 計 | 428 | 計 | 1165 | 計 | 959 |

表 3.5: 多義性解消の結果

| | 語義数 | 実験文数 | 正解 | 誤り | 解答なし | 解答率 | 正解率 |
|---|---|---|---|---|---|---|---|
| 足跡 | 3 | 35 | 13 | 15 | 7 | 80.0 | 46.4 |
| 雨 | 3 | 165 | 91 | 30 | 44 | 73.3 | 75.2 |
| キャンプ | 4 | 30 | 11 | 4 | 15 | 50.0 | 73.3 |
| たばこ | 2 | 107 | 41 | 6 | 60 | 43.9 | 87.2 |
| 隣 | 2 | 45 | 9 | 6 | 30 | 33.3 | 60.0 |
| 波 | 4 | 56 | 3 | 20 | 33 | 41.1 | 13.0 |
| 日 | 6 | 84 | 6 | 44 | 34 | 59.5 | 12.0 |
| 光 | 2 | 114 | 30 | 25 | 59 | 48.2 | 54.5 |
| 昼 | 3 | 39 | 23 | 6 | 10 | 74.4 | 79.3 |
| 歴史 | 3 | 100 | 22 | 23 | 55 | 45.0 | 48.8 |
| 計 | - | 775 | 249 | 179 | 347 | 55.2 | 58.2 |

表 3.6: シソーラスを用いた多義性解消

| | 語義数 | 実験文数 | 正解 | 誤り | 解答なし | 正解率 |
|---|---|---|---|---|---|---|
| 足跡 | 3 | 7 | 1 | 2 | 4 | 33.3 |
| 雨 | 3 | 44 | 11 | 25 | 8 | 30.6 |
| キャンプ | 4 | 15 | 0 | 3 | 12 | 0.0 |
| たばこ | 2 | 60 | 14 | 24 | 22 | 36.8 |
| 隣 | 2 | 30 | 14 | 13 | 3 | 51.9 |
| 波 | 4 | 33 | 5 | 16 | 12 | 23.8 |
| 日 | 6 | 34 | 3 | 16 | 15 | 15.8 |
| 光 | 2 | 59 | 25 | 26 | 8 | 49.0 |
| 昼 | 3 | 10 | 2 | 3 | 5 | 40.0 |
| 歴史 | 3 | 55 | 13 | 8 | 34 | 61.9 |
| 計 | - | 347 | 88 | 136 | 123 | 39.3 |

注) プログラムの関係上全ての実験文に対して解を与えることはできなかった.